\documentclass[11pt, a4paper]{article}

\usepackage[utf8]{inputenc}
\usepackage[T1]{fontenc}
\usepackage{acl2017}
\usepackage{subfig}
\usepackage{times}
\usepackage{latexsym}
\usepackage{amsmath, amssymb}
\usepackage{graphicx}
\usepackage{linguex}
\usepackage{dblfloatfix}
\newcommand{\depmem}[1] {\textbf{#1}}

\newcommand{\attractor}[1] {\underline{#1}}

\aclfinalcopy

\author{Émile Enguehard\textsuperscript{1} \qquad Yoav Goldberg\textsuperscript{2} \qquad Tal Linzen\textsuperscript{3,4}\\
    \textsuperscript{1}Département d'informatique, ENS, PSL Research University\\
    \textsuperscript{2}Computer Science Department, Bar Ilan University\\
    \textsuperscript{3}LSCP \& IJN, CNRS, EHESS and ENS, PSL Research University\\
    \textsuperscript{4}Department of Cognitive Science, Johns Hopkins University\\
    \texttt{\{emile.enguehard,tal.linzen\}@ens.fr \quad yoav.goldberg@gmail.com}
}

\title{Exploring the Syntactic Abilities of RNNs with Multi-task Learning}

\begin{document}

\maketitle

\begin{abstract}
Recent work has explored the syntactic abilities of RNNs using the subject-verb agreement task, which diagnoses sensitivity to sentence structure. RNNs performed this task well in common cases, but faltered in complex sentences \cite{linzen2016assessing}. We test whether these errors are due to inherent limitations of the architecture or to the relatively indirect supervision provided by most agreement dependencies in a corpus. We trained a single RNN to perform both the agreement task and an additional task, either CCG supertagging or language modeling. Multi-task training led to significantly lower error rates, in particular on complex sentences, suggesting that RNNs have the ability to evolve more sophisticated syntactic representations than shown before. We also show that easily available agreement training data can improve performance on other syntactic tasks, in particular when only a limited amount of training data is available for those tasks. The multi-task paradigm can also be leveraged to inject grammatical knowledge into language models.
\end{abstract}

\section{Introduction}

Recurrent neural networks (RNNs) have seen rapid adoption in natural language processing applications. Since these models are not equipped with explicit linguistic representations such as dependency parses or logical forms, new methods are needed to characterize the linguistic generalizations that they capture. One such method is drawn from behavioral psychology: the network is tested on cases that are carefully selected to be informative as to the generalizations that the network has acquired.

\newcite{linzen2016assessing} have recently applied this methodology to evaluate how well a trained RNN captures sentence structure, using the \textbf{agreement prediction task} \cite{bock1991broken,elman1991distributed}. The form of an English verb often depends on its subject. Identifying the subject of a given verb of requires sensitivity to sentence structure. Consequently, testing an RNN on its ability to choose the correct form of a verb in context can shed light on the sophistication of its syntactic representations (see Section \ref{sec:agreement_prediction} for details).

RNNs trained specifically to perform the agreement task can achieve very good average performance on a corpus, with accuracy close to 99\%. However, error rates increase substantially on complex sentences \cite{linzen2016assessing,linzen2017agreement}, suggesting that the syntactic knowledge acquired by the RNN is imperfect. Finally, when the RNN is trained as a language model rather than specifically on the agreement task, its sensitivity to subject-verb agreement, measured as the relative probability of the grammatical and ungrammatical forms of the verb, degrades dramatically.

Are the limitations that RNNs showed in previous work inherent to their architecture, or can these limitations be mitigated by stronger supervision? We address this question using multi-task learning, where the same model is encouraged to develop representations that are simultaneously useful for multiple tasks. To provide the RNN with an incentive to develop more sophisticated representations, we trained it to perform one of two tasks: the first is combinatory categorical grammar (CCG) supertagging \cite{bangalore1999supertagging}, a sequence labeling task likely to require robust syntactic representations; the second task is language modeling.

We also investigate the inverse question: can tasks such as supertagging benefit from joint training with the agreement task? This question is of practical interest. Large training sets for the agreement task are much easier to create than training sets for supertagging, which are based on manually parsed sentences. If the training signal from the agreement prediction task proves to be beneficial for supertagging, this could lead to improved supertagging (and therefore parsing) performance in languages in which we only have a small amount of parsed training sentences.

We found that multi-task learning, either with LM or with CCG supertagging, improved the performance of the RNN on the agreement prediction task. The benefits of combined training with supertagging can be quite large: accuracy in challenging relative clause sentences increased from 50.6\% to 76.2\%. This suggests that RNNs are in principle capable of acquiring much better syntactic representations than those they learned from the corpus in \newcite{linzen2016assessing}. 

In the other direction, joint training on the agreement prediction task did not improve overall language model perplexity, but made the model more syntax-aware: grammatically appropriate verb forms had higher probability than grammatically inappropriate ones. When a limited amount of CCG training data was available, joint training on agreement prediction led to improved supertagging accuracy. These findings suggest that multi-task training with auxiliary syntactic tasks such as agreement prediction can lead to improved performance on standard NLP tasks.

\section{Background and Related Work}

\subsection{Agreement Prediction}
\label{sec:agreement_prediction}

English present-tense third-person verbs agree in number with their subject: singular subjects require singular verbs (\textit{the boy smiles}) and plural subjects require plural verbs (\textit{the boys smile}). Subjects in English are not overtly marked, and complex sentences often have multiple subjects corresponding to different verbs. Identifying the subject of a particular verb can therefore be non-trivial in sentences that have multiple nouns:

\ex.The only \attractor{championship} \depmem{banners} that are currently displayed within the \attractor{building} \depmem{are} for national or NCAA Championships.

Determining that the subject of the verb in boldface is \textit{banners} rather than the singular nouns \textit{championship} and \textit{building} requires an understanding of the structure of the sentence.

In the agreement task, the learner is given the words leading up to a verb (a ``preamble''), and is instructed to predict whether that verb will take the plural or singular form. This task is modeled after a standard psycholinguistic task, which is used to study syntactic representations in humans \cite{bock1991broken,franck2002subject,staub2009interpretation,bock2011reaching}.

Any English sentence with a third-person present-tense verb can be used as a training example for this task: all we need is a tagger that can identify such verbs and determine whether they are plural or singular. As such, large amounts of training data for this task can be obtained from a corpus.

The agreement task can often be solved using simple heuristics, such as copying the number of the most recent noun. It can therefore be useful to evaluate the model using sentences in which such a heuristic would fail because one or more nouns of the opposite number from the subject intervene between the subject and the verb; such nouns ``attract'' the agreement away from the grammatical subject. In general, the more such attractors there are the more difficult the task is for a sequence model that does not represent syntax (we focus on sentences in which \textbf{all} of the nouns between the subject and the verb are of the opposite number from the subject):

\ex.The \depmem{number} of \attractor{men} \depmem{is} not clear. (One attractor)

\ex.The \depmem{ratio} of \attractor{men} to \attractor{women} \depmem{is} not clear. (Two attractors)

\ex.The \depmem{ratio} of \attractor{men} to \attractor{women} and \attractor{children} \depmem{is} not clear. (Three attractors)

\subsection{CCG Supertagging}

Combinatory Categorial Grammar (CCG) is a syntactic formalism that relies on a large inventory of lexical categories \cite{steedman2000syntactic}. These categories are known as \emph{supertags}, and can be thought of as a fine-grained extension of the usual part-of-speech tags. For example, intransitive verbs (\textit{smile)}, transitive verbs (\textit{build}) and raising verbs (\textit{seem}) all have different tags: \textit{S\textbackslash NP}, \textit{(S\textbackslash NP)/NP} and \textit{(S\textbackslash NP)/(S\textbackslash NP)}, respectively.

CCG parsers typically rely on a supertagging step where each word in a sentence is associated with an appropriate tag. In fact, supertagging is almost as difficult as finding the full CCG parse of the sentence: once the supertags are determined, only a small number of parses are possible. At the same time, supertagging is simple to set up as a machine learning problem, since at each word it amounts to a straightforward classification problem \cite{bangalore1999supertagging}. RNNs have shown excellent performance on this task, at least in English \cite{xu2015ccg,lewis2016lstm,vaswani2016supertagging}.

In contrast with the agreement task, training data for supertagging needs to be obtained from parsed sentences which require expert annotation \cite{hockenmaier2007ccgbank}; the amount of training data is therefore limited even in English, and much more sparse in other languages.

\subsection{Language Modeling}

The goal of a language model is to learn the distribution $\hat{p}(w_j | w_1,\dots,w_{j-1})$ of the $j$-th word in a sentence given the $j-1$ words preceding it. We seek to minimize the mean negative log-likelihood of all sentences $s_i = w_{i,1}\ldots{}w_{i,n_i}$ in our data:

\begin{multline}\label{eq:lmloss}
L(\hat{p}) = -\frac{1}{Z} \sum_{i=1}^N \sum_{j=1}^{n_i} \log \hat{p} (w_{i,j} | w_{i,1:j-1})
\end{multline}

where $Z = \sum_{i=1}^N n_i$. Language modeling performance is often quantified using the perplexity $2^{L(\hat{p})}$. The effectiveness of RNNs in language modeling, in particular LSTMs, has been demonstrated in numerous studies \cite{mikolov2010recurrent,sundermeyer2012lstm,jozefowicz2016exploring}.

\subsection{Multitask Learning}

The benefits of multi-task learning in neural networks are straightforward. Neural networks often require a large amount of training data to achieve good performance on a task. Even with a significant amount of training data, the signal may be too sparse for them to pick it up given their weak inductive biases. By training a network on a simple task for which large quantities of data are available, we can encourage it to evolve representations that would help its performance on the primary task \cite{caruana1998multitask,bakker2003task}. This logic has been applied to various NLP tasks, with generally encouraging results \cite{collobert2008unified,hashimoto2016joint,sogaard2016deep,alonso2016multitask,bingel2017identifying}.

\section{Methods}

\subsection{Datasets}

We used two training datasets. The first is the corpus of approximately $1.5$ million sentences from the English Wikipedia compiled by \newcite{linzen2016assessing}. All sentences had at most $50$ words and contained at least one third-person present-tense agreement dependency. Following \newcite{linzen2016assessing}, we replaced rare words by their part-of-speech tags, using the Penn Treebank tag set \cite{marcus1993building}.\footnote{In the LM experiments, we restricted ourselves to $10000$ words, amounting to $91.2\%$ of the all occurrences. In the CCG supertagging experiments, we used those $12,126$ words that occurred more than $150$ times, amounting to $92.2\%$ of the total number of occurrences.}

The second data set we used is the CCG-Bank \cite{hockenmaier2007ccgbank}, a CCG version of the Penn Treebank. This corpus contained $48934$ English sentences, $27299$ of which include a present tense third-person verb agreement dependency. A negligible number of sentences longer than $90$ words were removed. We applied the traditional split where Sections~2-21 are used for training and Section~23 for testing ($41294$ and $2407$ sentences respectively).\footnote{For experiments using this corpus, we use $15784$ words occurring at least four times, amounting to $95.9\%$ of occurrences, and replace other words by their POS tags.} Out of the $1363$ different supertags that occur in the corpus, we only attempted to predict the $452$ supertags that occurred at least ten times; we replaced the rest (0.2\% of the tokens) by a dummy value.

\subsection{Model}

The model in all of our experiments was a standard single-layer LSTM.\footnote{Our code and data are available at \url{https://github.com/emengd/multitask-agreement}.} The first layer was a vector embedding of word tokens into $D$-dimensional space. The second was a $D$-dimensional LSTM. The following layers depended on the task. For agreement, the output layers consisted of a linear layer with a one-dimensional output and a sigmoid activation; for language modeling, a linear layer with an $N$-dimensional output, where $N$ is the size of the lexicon, and a softmax activation; and for supertagging, a linear layer with an $S$-dimensional output, where $S$ is the number of possible tags, followed by a softmax activation.

 The language modeling loss is the mean negative log-likelihood of the data given in Equation~\eqref{eq:lmloss}; the loss for agreement is the mean binary cross-entropy of the classifier:
\begin{equation*}
L_\text{agr} = - \frac{1}{|S|} \sum_{s\in S} \log \left(\hat{q}(\text{num}(s) | s_{:\text{vb}})\right)
\end{equation*}
where $\hat{q}$ is the estimated distribution of verb numbers, $S$ the set of sentences, $\text{num}(s)$ the correct verb number in $s$ and $s_{:\text{vb}}$ the sentence up to the verb. The loss for CCG supertagging is the mean cross-entropy of the classifiers:
\begin{equation*}
L_\text{ST} = - \frac{1}{\sum_{s} |s|} \sum_{s \in S} \sum_{w_j \in s} \log \left(\hat{r}(\text{tag}(w_j) | s_{:w_j})\right)
\end{equation*}
where $\hat{r}$ is the estimated distribution of CCG supertags, $\text{tag}(w_j)$ is the correct tag of word~$w_j$ in~$s$, and $s_{:w_j}$ is the sentence~$s$ up to and including $w_j$.

We had at most two tasks in any given experiment. We considered two separate setups for learning from those two tasks: joint training and pre-training. 

\paragraph{Joint training:} 

In this setup we had parallel output layers for each task. Both output layers received the shared LSTM representations as their input. We define the global loss $L$ as follows:

\begin{equation}\label{eq:loss}
L = \frac{1}{1+r}L_1 + \frac{r}{1+r} L_2
\end{equation}

where $L_1$ and $L_2$ are the losses associated with each task, and $r$ is the weighting ratio of task 2 relative to task 1. This means that $r$ is a hyperparameter that needs to be tuned. Note that sample averaging occurs before formula~\eqref{eq:loss} is applied.

\paragraph{Pre-training:}

In this setup, we first trained the network on one of the tasks; we then used the weights learned by the network for the embedding layer and the LSTM layer as the initial weights of a new network which we then trained on the second task.

\subsection{Training}

All neural networks were implemented in Keras \cite{chollet2015keras} and Theano \cite{theano}. We use the AdaGrad optimizer. We use batch training with batch sizes 128 for language modeling experiments and 256 for supertagging experiments on supertagging.

\section{Agreement and Supertagging}

For the supertagging experiments we used the full CCG corpus as well as $30\%$ of the Wikipedia corpus for the agreement task ($20\%$ for training and $10\%$ for testing). We trained the model for 20 epochs. The accuracy figures we report are averaged across three runs. We set the size of the network $D$ to $500$ hidden units.\footnote{In initial experiments $D = 50$ yielded supertagging results inferior to a majority choice baseline.} We ran a single pre-training experiment in each direction, as well as four joint training experiments, with the weight $r$ of the agreement task set to $0.1$, $1$, $10$ or $100$. 

We considered two baselines for the agreement task: the last noun baseline predicts the number of the verb based on the number of the most recent noun, and the majority baseline always predicts a singular verb (singular verbs are more common than plural ones in our corpus). Our baseline for supertagging was a majority baseline that predicts for each word its most common supertag. 

The agreement task predicts the number of the verb based only on its left context (the preamble). We trained our supertagging model in the same setup. Since our model did not have access to the right context of a word when determining its supertag, we could not expect to compete with state-of-the-art taggers that use right-context lookahead \cite{xu2015ccg} or even bidirectional RNNs that read the entire sentence from right to left \cite{vaswani2016supertagging,lewis2016lstm}; we therefore did not compare our accuracy to these taggers.

\subsection{Overall Results}

\begin{figure}[bt]
\centering
\subfloat[][Agreement]{
    \includegraphics[width=\linewidth]{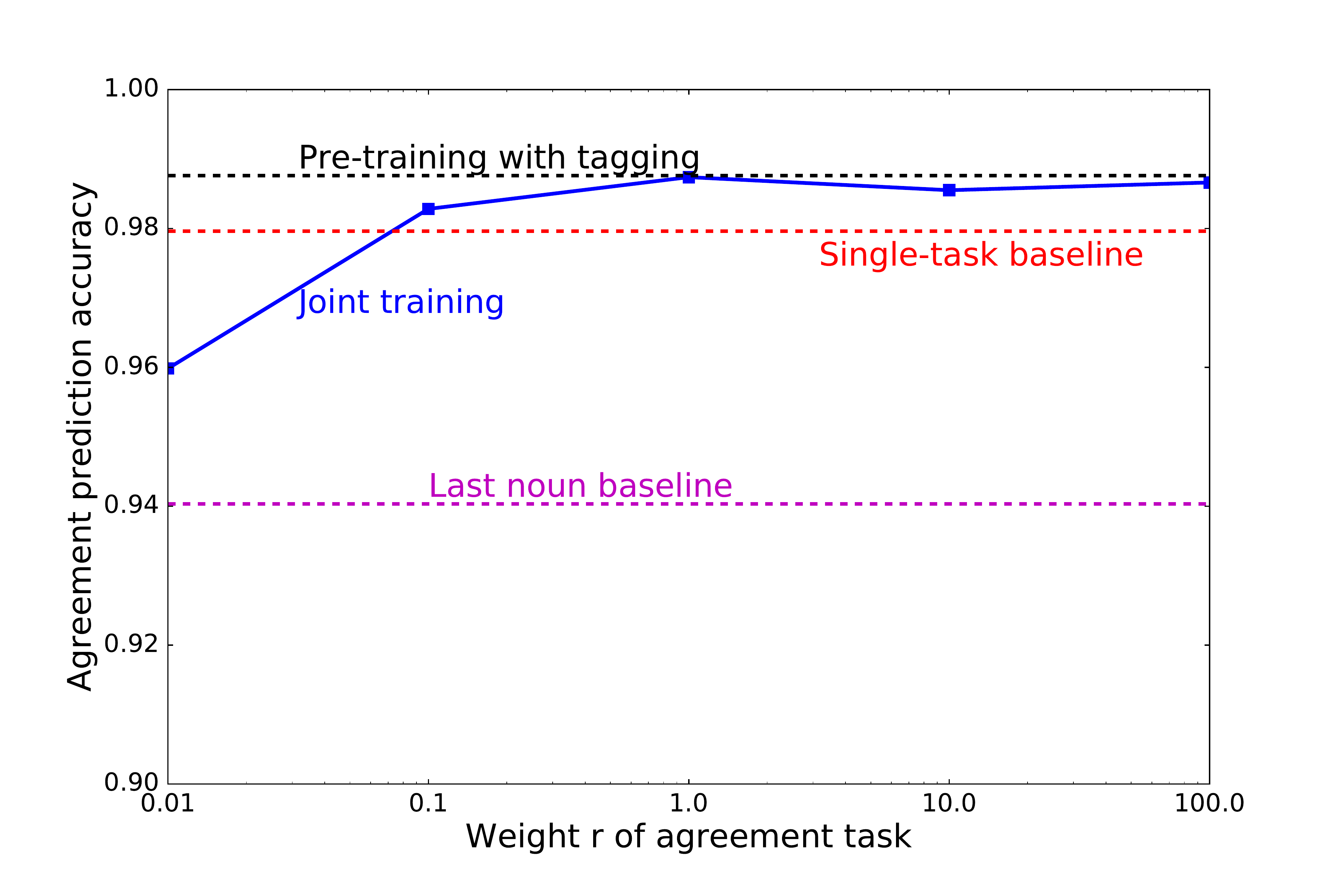}
    \label{fig:overall_st_agr}
}

\subfloat[][Supertagging]{
    \includegraphics[width=\linewidth]{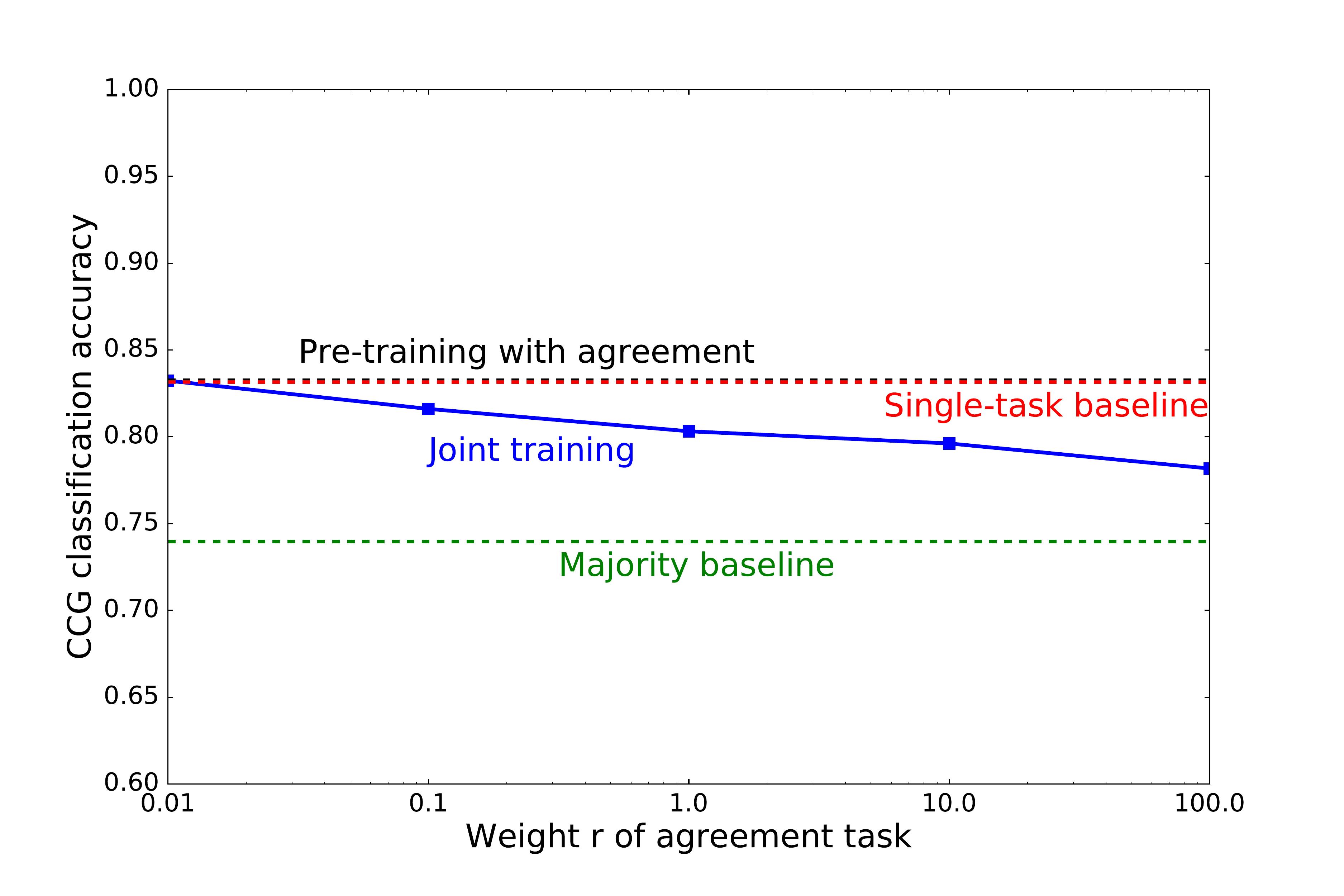}
    \label{fig:overall_st_st}
}
\caption{Overall results of supertagging + agreement multi-task training.}
\label{fig:bicorpus_tag}
\end{figure}

Figure \ref{fig:bicorpus_tag} shows the overall results of the experiment.
Multi-task training with supertagging significantly improved overall accuracy on the agreement task (Figure \ref{fig:overall_st_agr}), either with pre-training or joint training: compared to the single-task setup, the agreement error rate decreased by up to 40\% in relative terms (from 2.04\% to 1.24\%). Conversely, multi-task training with agreement did not improve supertagging accuracy, either in the pre-training or in the joint training regime; supertagging accuracy decreased the higher the weight of the agreement task (Figure \ref{fig:overall_st_st}). 

Comparing the two multi-task learning regimes, the pre-training setup performed about as well as the joint training setup with the optimal $r$. In the following supertagging experiments we dispensed with the joint training setup, which is time consuming since it requires trying multiple values of $r$, and focused only on the pre-training setup.

\subsection{Effect of Corpus Size}

To further investigate the relative contribution of the two supervision signals, we conducted a series of follow-up experiments in the pre-training setup, using subsets of varying size of both corpora. We also included POS tagging as an auxiliary task to determine to what extent the full parse of the sentence (approximated by supertags) is crucial to the improvements we have seen in the agreement task. Since POS tags contain less syntactic information than CCG supertags, we expect them to be less helpful as an auxiliary task. Penn Treebank POS tags distinguish singular and plural nouns and verbs, but CCG supertags do not; to put the two tasks on equal footing we removed number information from the POS tags.  We trained for 15~epochs and averaged our results over 5~runs.

\begin{figure}[bt]
\centering
\subfloat[][Agreement]{
    \includegraphics[width=0.95\linewidth]{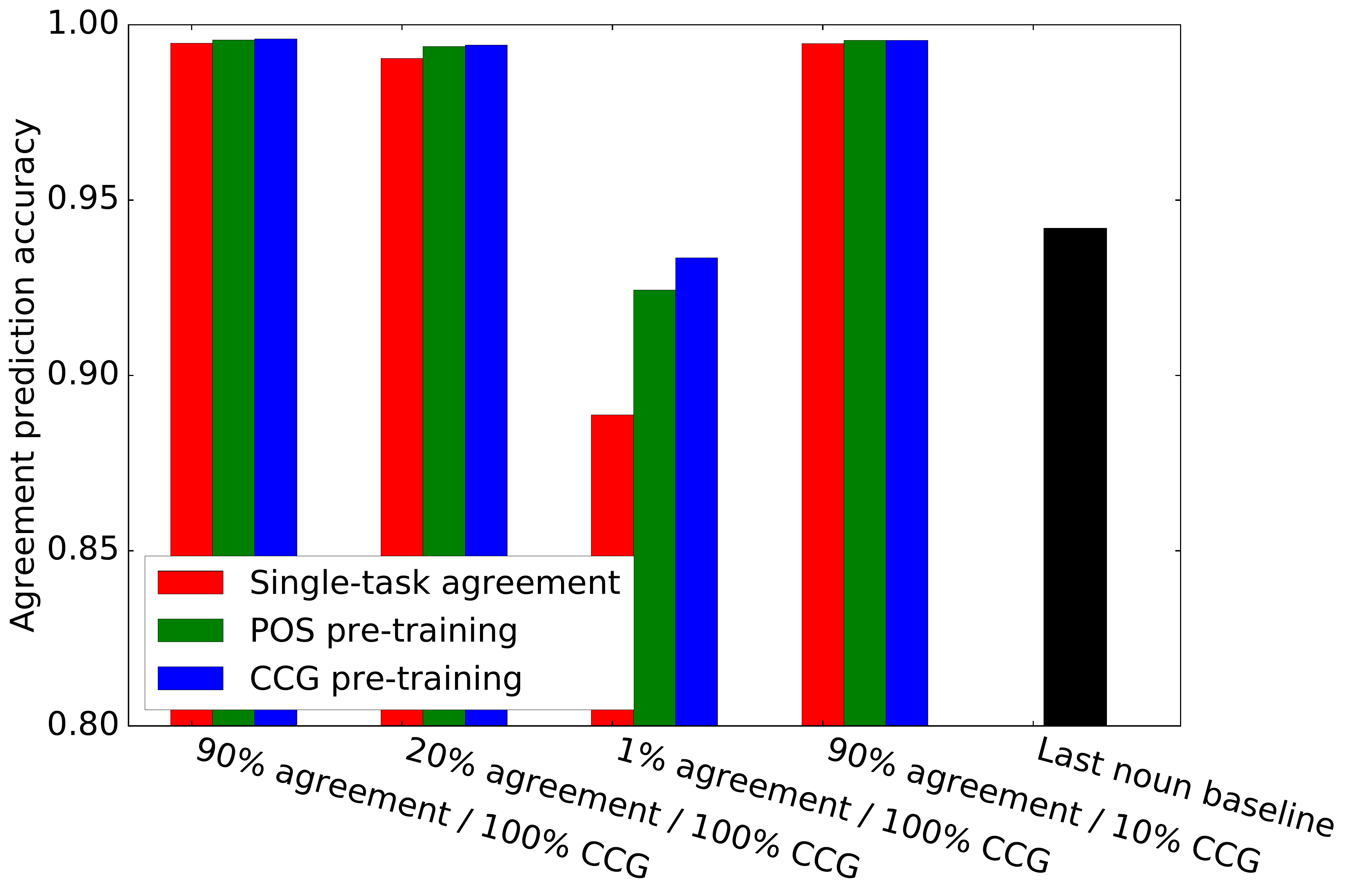}
    \label{fig:corpus_size_agreement}
}

\subfloat[][Supertagging]{
    \includegraphics[width=0.95\linewidth]{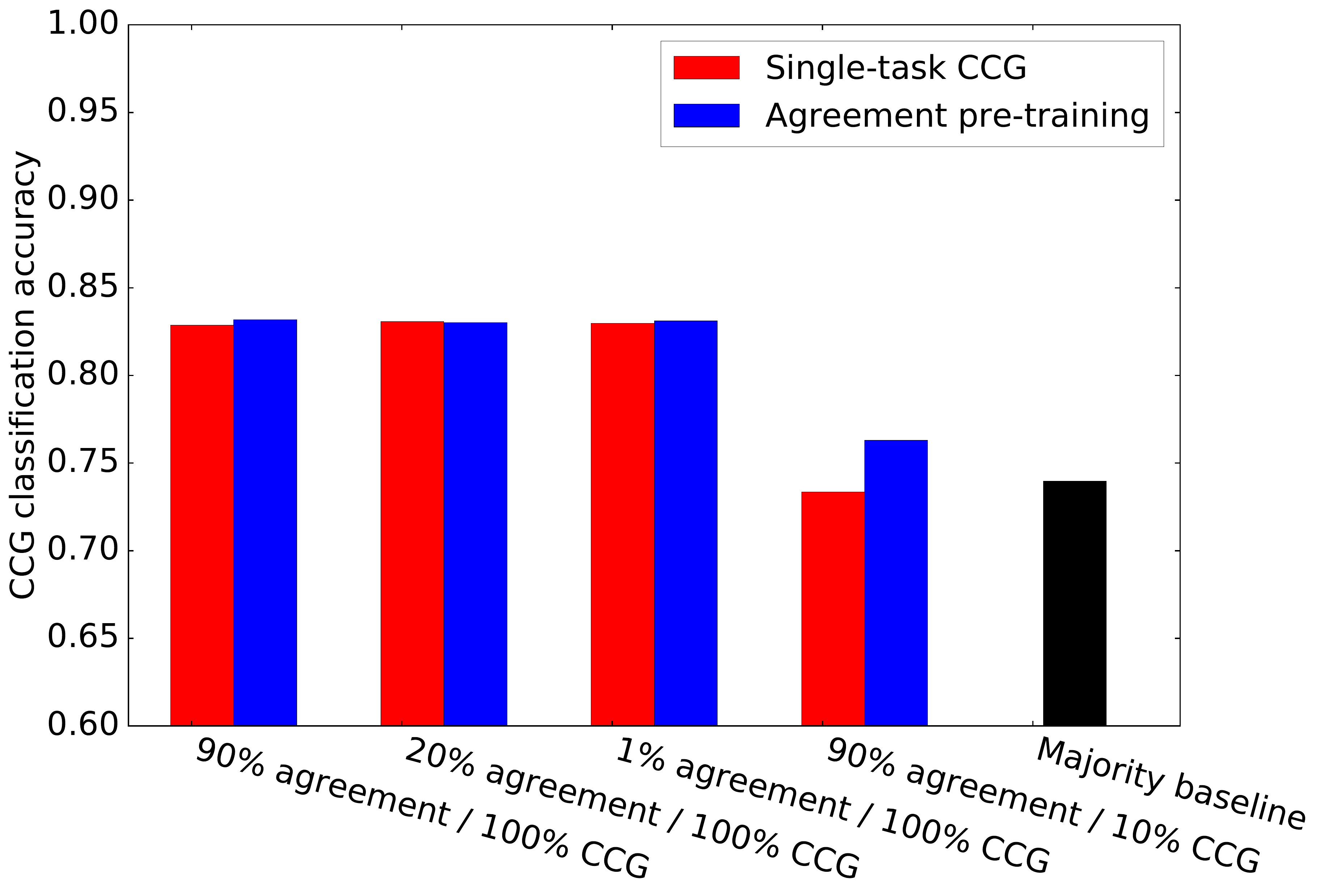}
    \label{fig:corpus_size_st}
}
\caption{The effect of corpus size on agreement and supertagging accuracy in multi-task settings.}
\label{fig:size_agr}
\end{figure}

The results for the agreement task are shown in Figure~\ref{fig:corpus_size_agreement} (baseline values are always calculated over the full corpora). The figure confirms the beneficial effect of supertagging pre-training (note that the scale starts at $0.8$, not $0.9$ as in Figure~\ref{fig:overall_st_agr}). This effect was amplified when we used less training data for the agreement task. Pre-training on POS tagging yielded a similar though slightly weaker effect. This suggests that much of the improvement in syntactic representations due to pre-training on supertagging can also be gained from pre-training on POS tagging.

Finally, Figure~\ref{fig:corpus_size_st} shows that pre-training on the agreement task improved supertagging accuracy when we only used 10\% of the CCG corpus (increase in accuracy from 73.4\% to 76.3\%); however, even with agreement pre-training supertagging accuracy is lower than when the model is trained on the full CCG corpus (where accuracy was 83.1\%).

In summary, the data for each task can be used to supplement the data for the other, but there is a large imbalance in the amount of information provided by each task. This is not surprising given that the CCG supertagging data is much richer than the agreement data for any individual sentence. Still, we showed that the syntactic signal from the agreement prediction task can help improve parsing performance when CCG training data is sparse; this weak but widely available source of syntactic supervision may therefore have a practical use in languages with smaller treebanks than English.

\subsection{Attraction Errors}

Most sentences are syntactically simple and do not pose particular challenges to the models: the accuracy of the last noun baseline in Figure \ref{fig:overall_st_agr} was close to 95\%. To investigate the behavior of the model on more difficult sentences, we next break down our test sentences by the number of agreement attractors (see Section \ref{sec:agreement_prediction}).

\begin{figure}[bt]
\centering
\includegraphics[width=1.05\linewidth]{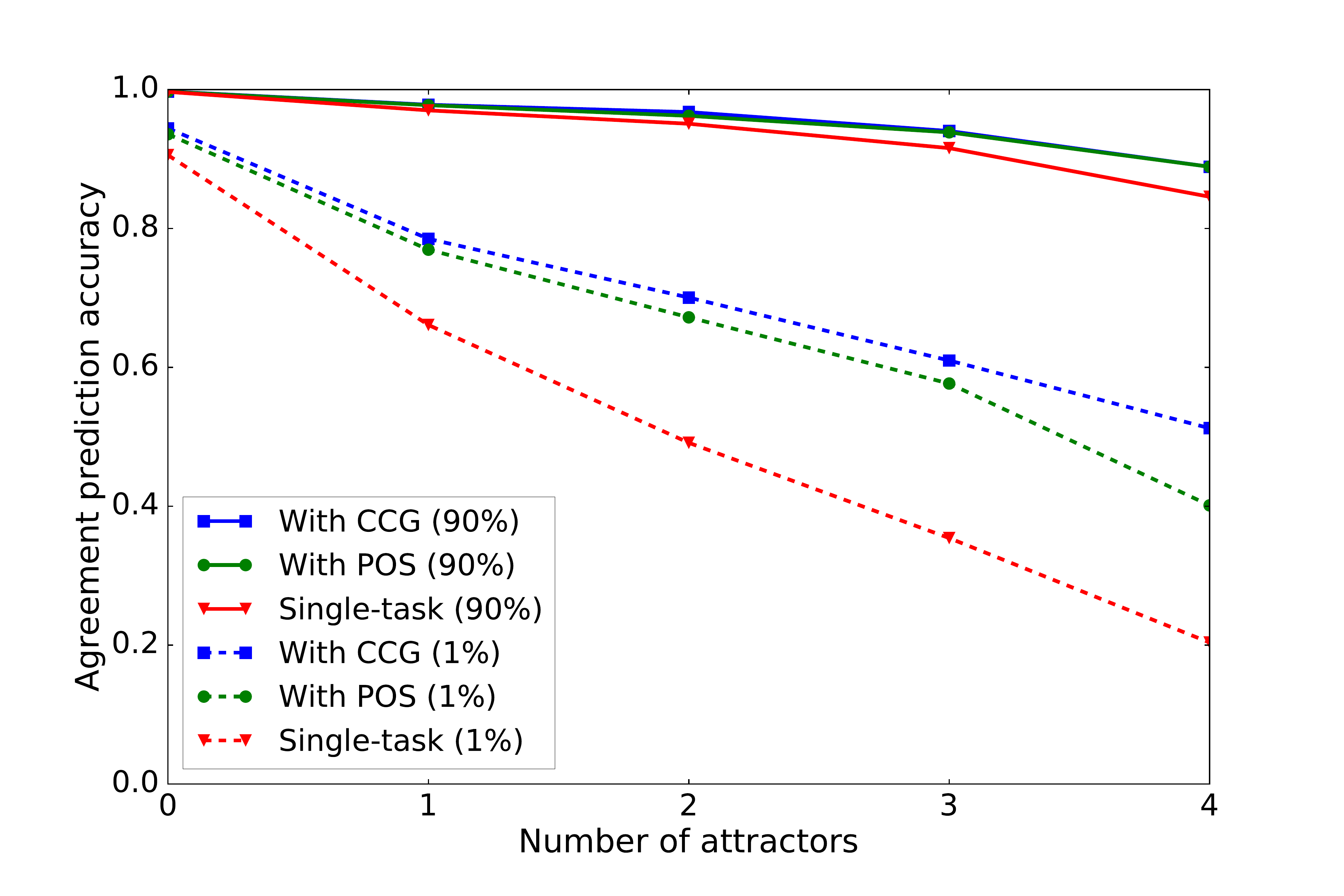}
\caption{Agreement accuracy as a function of the number of attractors intervening between the subject and the verb, for two different subsets of the agreement corpus (90\% and 1\% of the corpus).}
\label{fig:size_ndi}
\end{figure}

Our results, shown in Figure~\ref{fig:size_ndi}, confirm that attractors make the agreement task more difficult, and that pre-training helps overcome this difficulty. This effect is amplified when we only use a small subset of the agreement corpus. In this scenario, the accuracy of the single-task model on sentences with four attractors is only 20.4\%. Pre-training makes it possible to overcome this difficulty to a significant extent (though not entirely), increasing the accuracy to 40.1\% in the case of POS tagging and 51.2\% in the case of supertagging. This suggests that a network that has developed sophisticated syntactic representations can transfer its knowledge to a new syntactic task using only a moderate amount of data.

\begin{figure}[h!]
\centering
    \includegraphics[width=\linewidth]{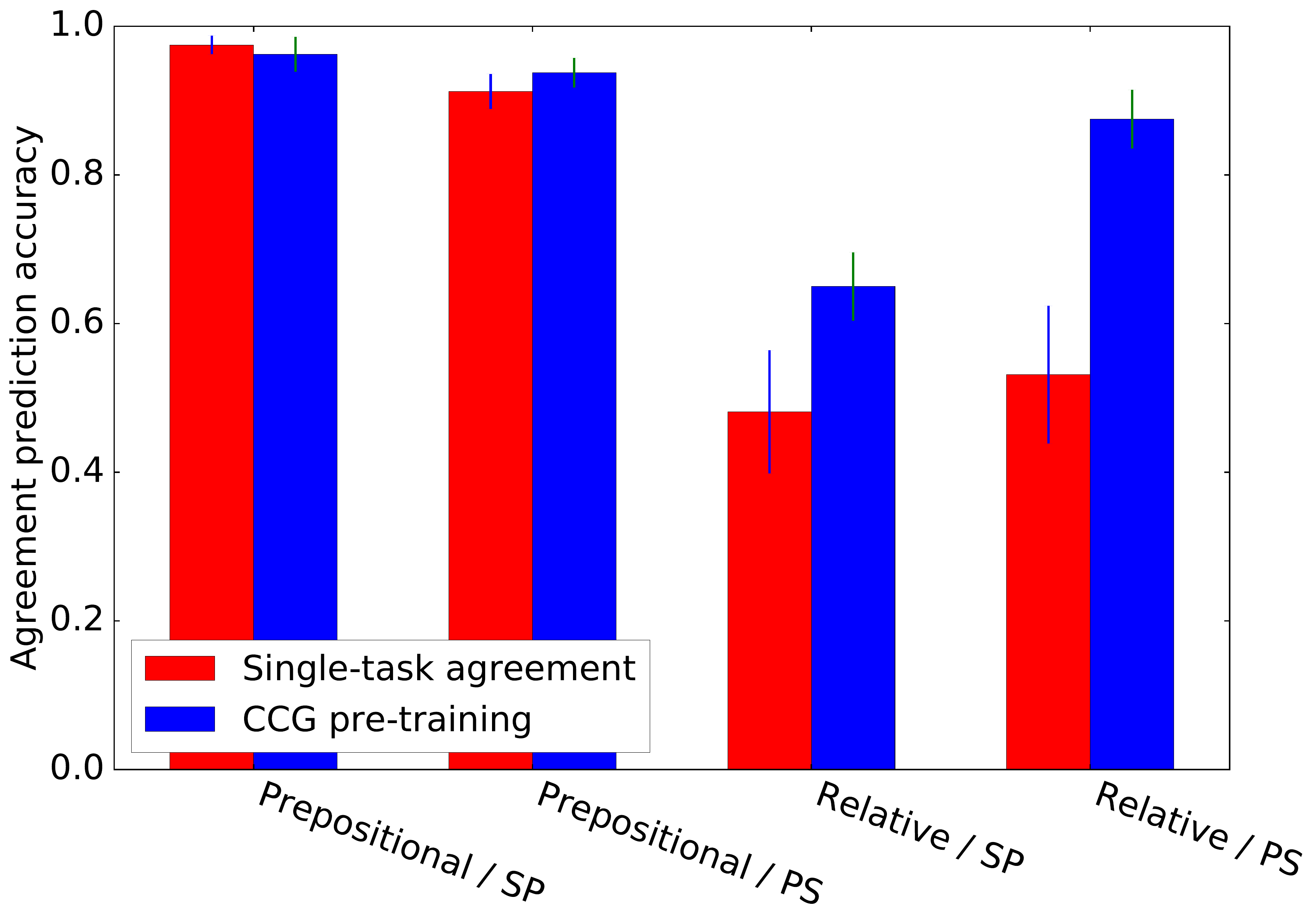}

\caption{Accuracy on sentences from \newcite{bock1992regulating}.  Error bars indicate standard deviation across runs.}
    \label{fig:bock_results}
\end{figure}

\subsection{Relative Clauses}
\label{sec:relative_clauses}

In \newcite{linzen2016assessing}, attraction errors were particularly severe when the attractor was inside a relative clause. To gain a more precise understanding of the errors and the extent to which pre-training can mitigate them, we turn to two sets of carefully constructed sentences from the psycholinguistic literature \cite{linzen2017agreement}. \newcite{bock1992regulating} compared preambles with prepositional phrase modifiers to closely matched relative clause modifiers:

\ex.\textsc{Prepositional:} The demo tape(s) from the popular rock singer(s)...
    
\ex.\textsc{Relative:} The demo tape(s) that promoted the popular rock singer(s)...

They constructed 24 such sentence pairs. Each of the sentences in each pair has four versions, with all possible combinations of the number of the subject and the attractor. We refer to them as SS for singular-singular (\textit{tape}, \textit{singer}), SP for singular-plural (\textit{tape}, \textit{singers}), and likewise PS and PP. We replaced out-of-vocabulary words with their POS, and further streamlined the materials by always using \textit{that} as the relativizer.

We retrained the single-task and pre-trained models on 90\% of the Wikipedia corpus. Like humans, neither model had any issues with SS and PP sentences, which do not have an attractor. The results for SP and PS sentences are shown in Figure \ref{fig:bock_results}. The comparison between prepositional and relative modifiers shows that the single-task model was much more likely to make errors when the attractor was in a relative clause (whereas humans are not sensitive to this distinction). This asymmetry was substantially mitigated, though not completely eliminated, by CCG pre-training.

Our second set of sentences was based on the experimental materials of \newcite{wagers2009agreement}. We adapted them by deleting the relativizer and creating two preambles from each sentence in the original experiment:

\ex.\textsc{Embedded verb:} The player(s) the coach(es)...

\ex.\textsc{Main clause verb:} The player(s) the coach(es) like the best...

In the first preamble, the verb is expected to agree with the embedded clause subject (\textit{the coach(es)}), whereas in the second one it is expected to agree with the main clause subject (\textit{the player(s)}). 

\begin{figure}[b]
    \centering
    \includegraphics[width=0.9\linewidth]{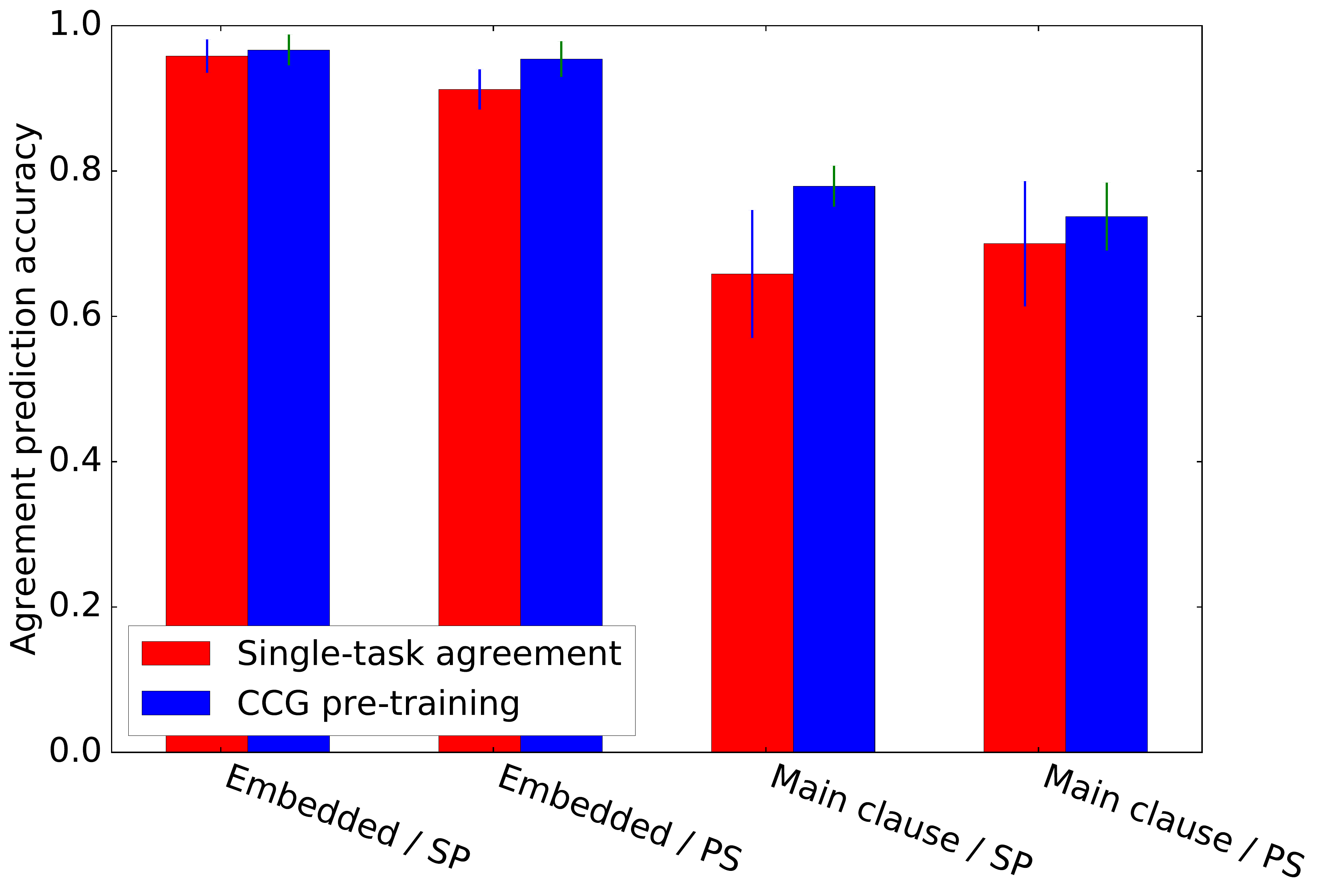}
\caption{\label{fig:wagers_results}Accuracy on sentences based on \newcite{wagers2009agreement}. Error bars indicate standard deviation across runs.}
\end{figure}

Figure \ref{fig:wagers_results} shows that both models made very few errors predicting the embedded clause verb, and more errors predicting the main clause verb. The relative improvement of the pre-trained model compared to the single-task one is more modest in these sentences, possibly because the single-task model does better to begin with on these sentences than on the \newcite{bock1992regulating} ones. This in turn may be because the attractor immediately precedes the verb in \newcite{bock1992regulating} but not in \newcite{wagers2009agreement}, and an immediately adjacent noun may be a stronger attractor. The Appendix contains additional figures tracking the predictions of the network as it processes a sample of sentences with relative clauses; it also illustrates the activation of particular units over the course of such a sentence.

\section{Agreement and Language Modeling}

We now turn our attention to the language modeling task. The previous experiments confirmed that agreement in sentences without attractors is easy to predict. We therefore limited ourselves in the language modeling experiments to sentences with potential attractors. Concretely, within the subset of 30\% of the Wikipedia corpus, we trained our language model only on sentences with at least one noun (of any number) between the subject and the verb. There were $60680$ sentences in the training set.  We averaged our results over three runs. Training was stopped after 10 epochs, and the number of hidden units was set to $D = 50$.

\subsection{Overall Results}

\begin{figure}[t]
\centering
\subfloat[][Agreement]{\includegraphics[width=\linewidth]{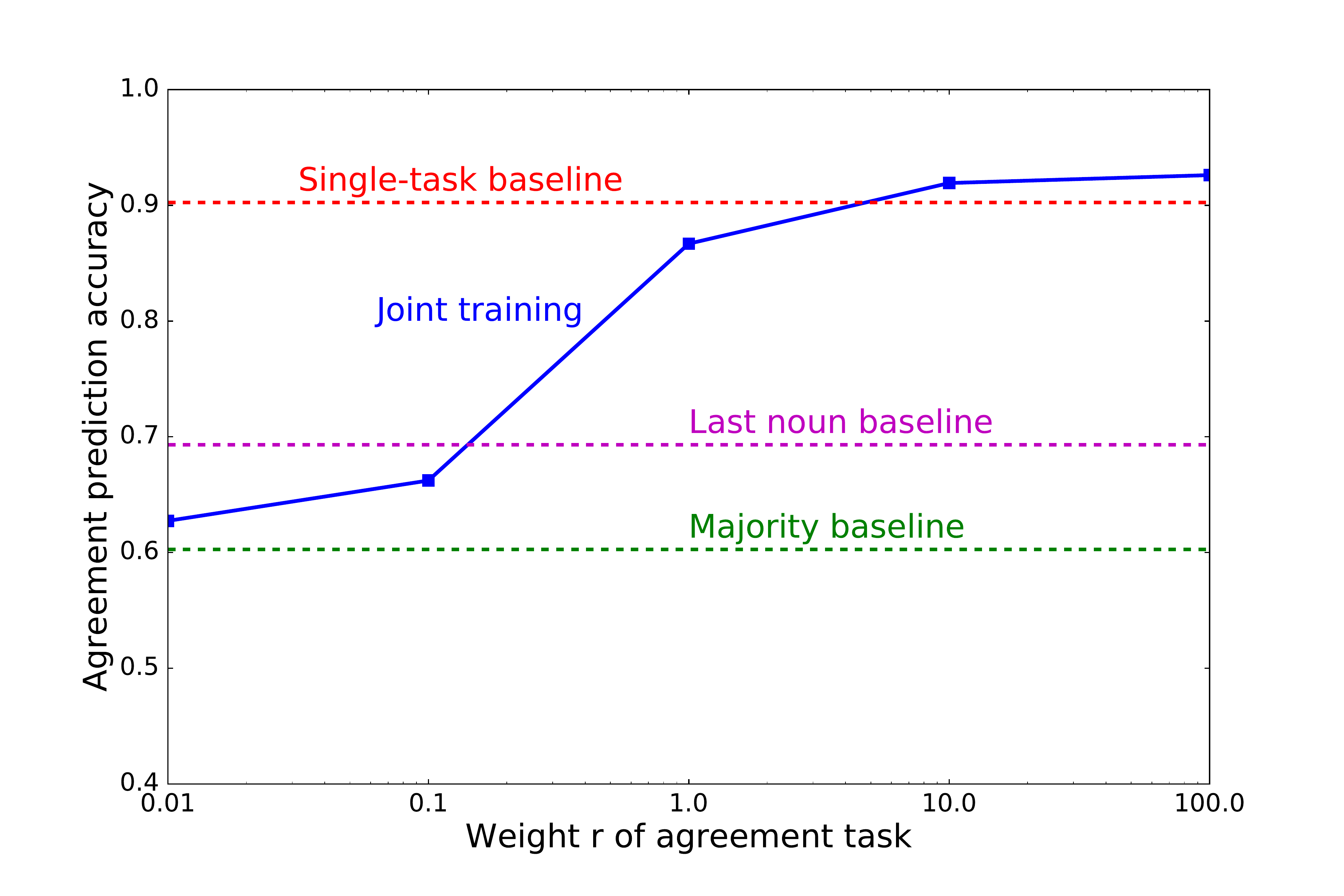}}

\subfloat[][Language modeling]{
    \includegraphics[width=0.95\linewidth]{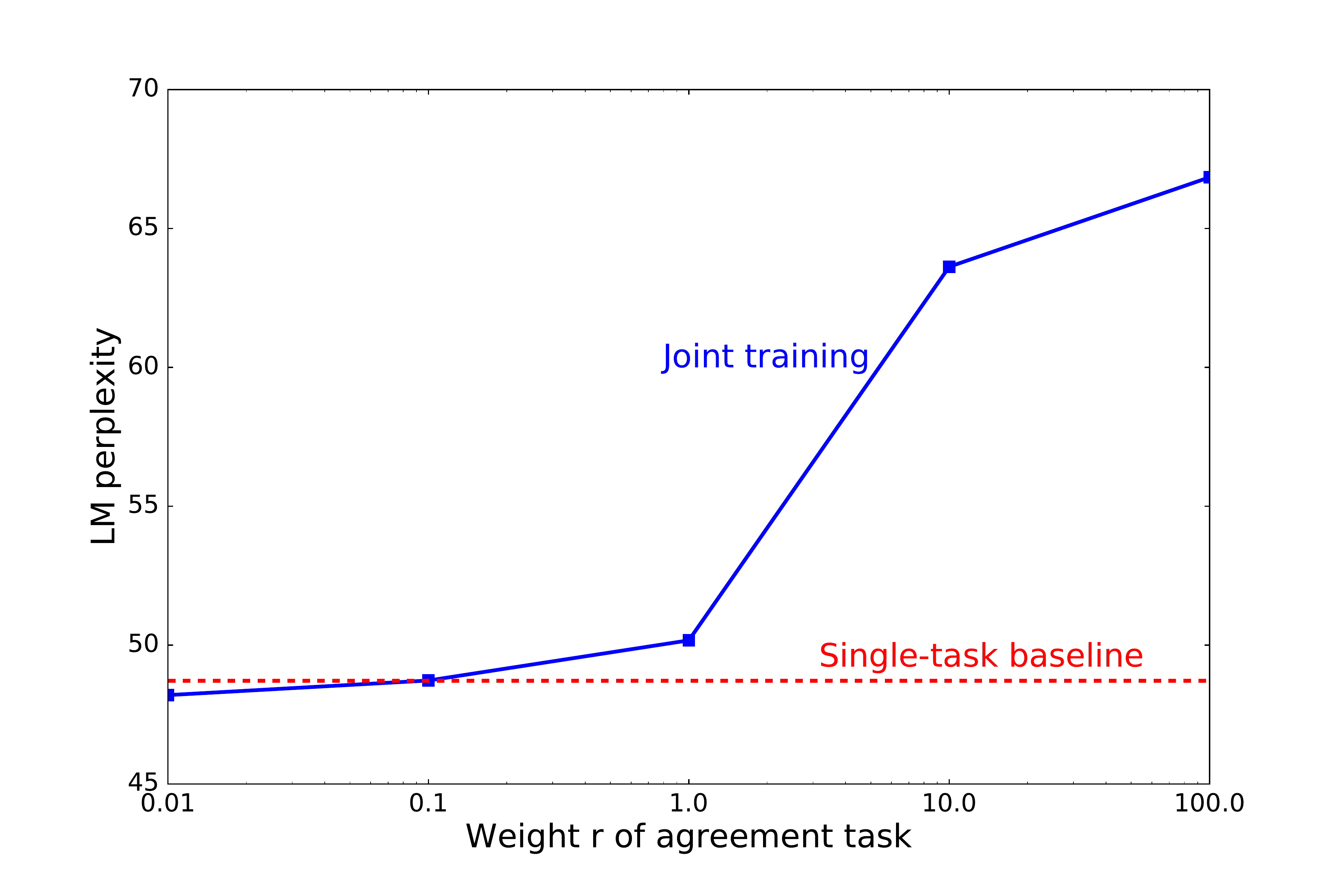}
    \label{fig:overall_lm_lm}
}
\caption{Overall results of language modeling~+ agreement multi-task training (trained only on sentences with an intervening noun).}
\label{fig:lmresults}
\end{figure}

The overall results are shown in Figure \ref{fig:lmresults}. Joint training with the LM task improves the performance of the agreement task to a significant extent, bringing accuracy up from 90.2\% to 92.6\% (a relative reduction of 25\% in error rate). This may be due to the higher quality of the word representations that can be learned from the language modeling signal, which in turn help the model make more accurate syntactic predictions.

In the other direction, we do not obtain clear improvements in perplexity from jointly training the LM with agreement. Surprisingly, visual inspection of Figure~\ref{fig:overall_lm_lm} suggests that the jointly trained LM may achieve somewhat better performance than the single-task baseline for \textit{small} values of $r$ (that is, when the agreement task has a small effect on the overall training loss). To assess the statistical significance of this difference, we repeated the experiment with $r = 0.01$ with 20 random initializations. The standard deviation in LM loss was about $0.018$, yielding a standard deviation of $0.011$ for three-run averages under Gaussian assumptions. Since the difference of $0.015$ between the mean LM losses of the single-task and joint training setups is of comparable magnitude, we conclude that there is no clear evidence that joint training reduces perplexity.

\subsection{Grammaticality of LM Predictions}

To evaluate the syntactic abilities of an RNN trained as a language model, \newcite{linzen2016assessing} proposed to perform the agreement task by comparing the probability under the learned LM of the correct and incorrect verb forms, under the assumption that all other things being equal a grammatical sequence should have a higher probability than an ungrammatical one \cite{lau2016grammaticality,legodais2017comparing}. For instance, if the sentence starts with \textit{the dogs}, we compute:

\begin{equation}\label{eq:thedogs}
\hat{p}_\text{correct} = \frac{\hat{p}(w_2 = \text{\small are} | w_{0:1} = \text{\small the dogs})}{\hat{p}(w_2 = \text{\small are} | \dots) + \hat{p}(w_2 = \text{\small is} | \dots)}
\end{equation}

\noindent The prediction for the agreement task is derived by thresholding $\hat{p}_\text{correct}$ at $0.5$. 

Is the LM learned in the joint training setup with high $r$ more aware of subject-verb agreement than a single-task LM? Note that this is not a circular question: we are not asking whether the explicit agreement prediction output layer can perform the agreement task --- that would be unsurprising --- but whether joint training with this task rearranges the probability distributions that the LM defines over the entire vocabulary in a way that is more consistent with English grammar.

As the method outlined in Equation \ref{eq:thedogs} may be sensitive to the idiosyncrasies of the particular verb being predicted, we also explored an unlexicalized way of performing the task. Recall that since we replace uncommon words by their POS tags, POS tags are part of our lexicon. We can use this fact to compare the LM probabilities of the POS tags for the correct and incorrect verb forms: in the example of the preamble \textit{the dogs}, the correct POS would be VBP and the incorrect one VBZ.

\begin{figure}[bt]
\centering
\includegraphics[width=\linewidth]{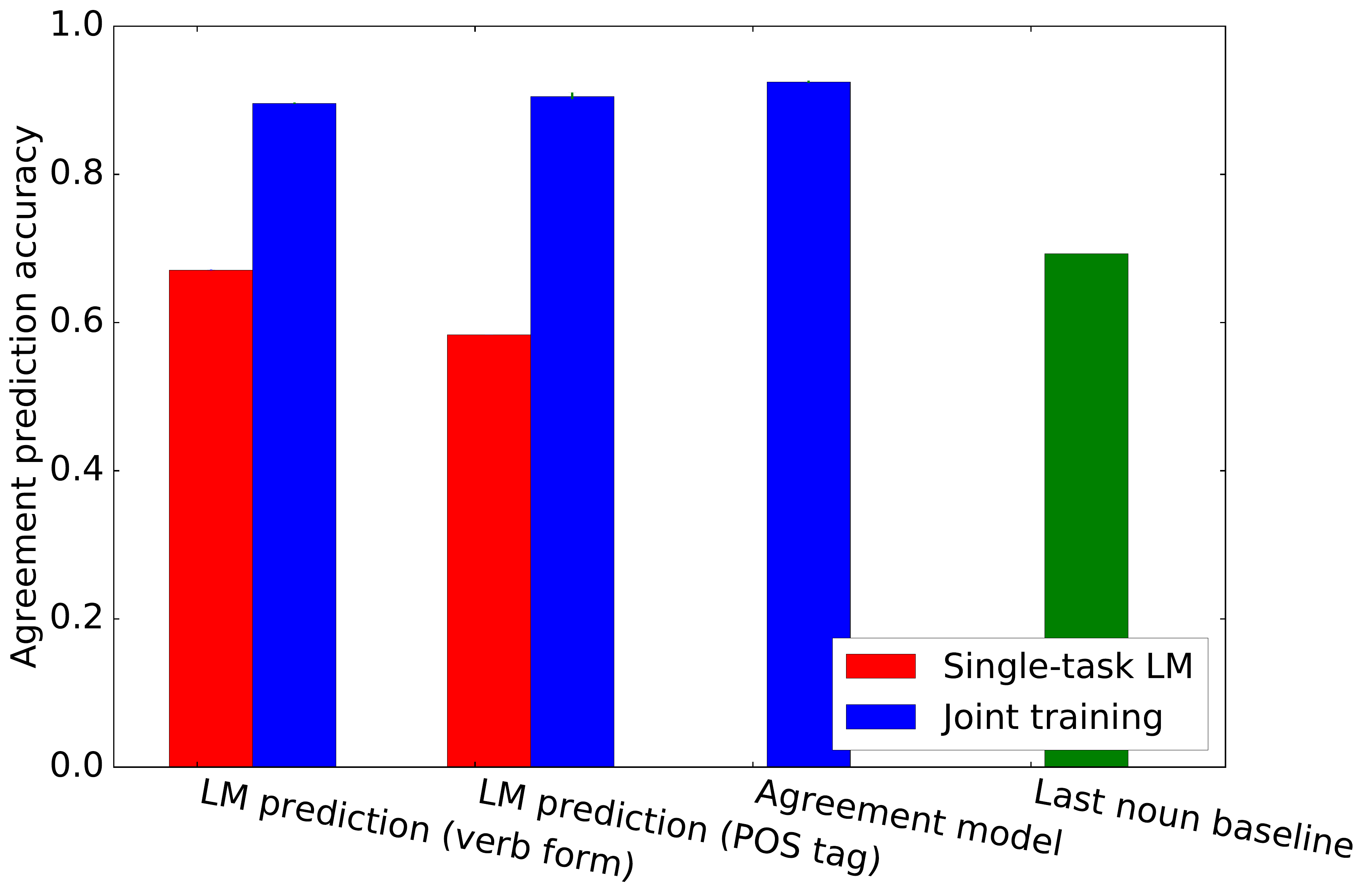}
\caption{Language model agreement evaluation. Red bars indicate the results obtained on the single-task LM model, blue bars those obtained in a joint training setup with $r = 100$.}
\label{fig:lmresults_preds}
\end{figure}

The results can be seen in Figure~\ref{fig:lmresults_preds}.  The accuracy of the LM predictions from the jointly trained models is almost as high as that obtained through the agreement model itself. Conversely, the single-task model trained only on language modeling performed only slightly better than chance, and worse than our last noun baseline (recall that the dataset only included sentences with an intervening noun between the subject and the verb, though possibly of the same number as the subject). Predictions based on POS tags are somewhat worse than predictions based on the specific verb. In summary, while joint training with the explicit agreement task does not noticeably reduce language model perplexity, it does help the LM capture syntactic dependencies: the ranking of upcoming words is more consistent with the constraints of English syntax.

\section{Conclusions}

Previous work has shown that the syntactic representations developed by RNNs that are trained on the agreement prediction task are sufficient for the majority of sentences, but break down in more complex sentences \cite{linzen2016assessing,linzen2017agreement}. These deficiencies could be due to fundamental limitations of the architecture, which can only be addressed by switching to more expressive architectures \cite{socher2014recursive,grefenstette2015learning,dyer2016recurrent}. Alternatively, they could be due to insufficient supervision signal in the agreement prediction task, for example because relative clauses with agreement attractors are infrequent in a natural corpus. 

We showed that additional supervision from pre-training on syntactic tagging tasks such as CCG supertagging can help the RNN develop more effective syntactic representations which substantially improve its performance on complex sentences, supporting the second hypothesis.

The syntactic representations developed by the RNNs were still not perfect even in the multi-task setting, suggesting that stronger inductive biases expressed as richer representational assumptions may lead to further improvements in syntactic performance. The weaker performance on complex sentences in the single-task setting indicates that the inductive bias inherent in RNNs is insufficient for learning adequate syntactic representations from unannotated strings; improvements due to a stronger inductive bias are therefore likely to be particularly pronounced in languages for which parsed corpora are small or unavailable. Finally, the strong syntactic supervision required to promote sophisticated syntactic representations in RNNs may limit their viability as models of language acquisition in children (though children may have sources of supervision that were not available to our models).

We also explored whether multi-task training with the agreement task can improve performance on more standard NLP tasks. We found that it can indeed lead to improved supertagging accuracy when there is a limited amount of training data for that task; this form of weak syntactic supervision can be used to improve parsers for low-resource languages for which only small treebanks are available. 

Finally, for language modeling, multi-task training with the agreement task did not reduce perplexity, but did improve the grammaticality of the predictions of the language model (as measured by the relative ranking of grammatical and ungrammatical verb forms); such a language model that favors grammatical sentences may produce more natural-sounding text.

\section*{Acknowledgments}
We thank Emmanuel Dupoux for discussion. This research was supported by the European Research Council (grant ERC-2011-AdG 295810 BOOTPHON), the Agence Nationale pour la Recherche (grants ANR-10-IDEX-0001-02 PSL and ANR-10-LABX-0087 IEC) and the Israeli Science Foundation (grant number 1555/15).

\bibliographystyle{acl_natbib}
\bibliography{rnn_multitask}

\begin{thebibliography}{}
\expandafter\ifx\csname natexlab\endcsname\relax\def\natexlab#1{#1}\fi

\bibitem[{Bakker and Heskes(2003)}]{bakker2003task}
Bart Bakker and Tom Heskes. 2003.
\newblock Task clustering and gating for {Bayesian} multitask learning.
\newblock {\em Journal of Machine Learning Research\/} 4:83--99.

\bibitem[{Bangalore and Joshi(1999)}]{bangalore1999supertagging}
Srinivas Bangalore and Aravind~K. Joshi. 1999.
\newblock Supertagging: An approach to almost parsing.
\newblock {\em Computational Linguistics\/} 25(2):237--265.

\bibitem[{Bingel and S{\o}gaard(2017)}]{bingel2017identifying}
Joachim Bingel and Anders S{\o}gaard. 2017.
\newblock Identifying beneficial task relations for multi-task learning in deep
  neural networks.
\newblock In {\em Proceedings of the 15th Conference of the European Chapter of
  the Association for Computational Linguistics: Volume 2, Short Papers\/}.
  Association for Computational Linguistics, Valencia, Spain, pages 164--169.

\bibitem[{Bock and Cutting(1992)}]{bock1992regulating}
Kathryn Bock and J.~Cooper Cutting. 1992.
\newblock Regulating mental energy: Performance units in language production.
\newblock {\em Journal of Memory and Language\/} 31(1):99--127.

\bibitem[{Bock and Middleton(2011)}]{bock2011reaching}
Kathryn Bock and Erica~L. Middleton. 2011.
\newblock Reaching agreement.
\newblock {\em Natural Language \& Linguistic Theory\/} 29(4):1033--1069.

\bibitem[{Bock and Miller(1991)}]{bock1991broken}
Kathryn Bock and Carol~A. Miller. 1991.
\newblock Broken agreement.
\newblock {\em Cognitive Psychology\/} 23(1):45--93.

\bibitem[{Caruana(1998)}]{caruana1998multitask}
Rich Caruana. 1998.
\newblock Multitask learning.
\newblock In Sebastian Thrun and Lorien Pratt, editors, {\em Learning to
  learn\/}, Kluwer Academic Publishers, Boston, pages 95--133.

\bibitem[{Chollet(2015)}]{chollet2015keras}
Fran\c{c}ois Chollet. 2015.
\newblock Keras.
\newblock \url{https://github.com/fchollet/keras}.

\bibitem[{Collobert and Weston(2008)}]{collobert2008unified}
Ronan Collobert and Jason Weston. 2008.
\newblock A unified architecture for natural language processing: Deep neural
  networks with multitask learning.
\newblock In {\em Proceedings of the 25th International Conference on Machine
  Learning\/}. New York, NY, USA, pages 160--167.

\bibitem[{Dyer et~al.(2016)Dyer, Kuncoro, Ballesteros, and
  Smith}]{dyer2016recurrent}
Chris Dyer, Adhiguna Kuncoro, Miguel Ballesteros, and A.~Noah Smith. 2016.
\newblock Recurrent neural network grammars.
\newblock In {\em {Proceedings of the 2016 Conference of the North American
  Chapter of the Association for Computational Linguistics: Human Language
  Technologies}\/}. Association for Computational Linguistics, pages 199--209.

\bibitem[{Elman(1991)}]{elman1991distributed}
Jeffrey~L. Elman. 1991.
\newblock Distributed representations, simple recurrent networks, and
  grammatical structure.
\newblock {\em Machine Learning\/} 7(2-3):195--225.

\bibitem[{Franck et~al.(2002)Franck, Vigliocco, and Nicol}]{franck2002subject}
Julie Franck, Gabriella Vigliocco, and Janet Nicol. 2002.
\newblock Subject-verb agreement errors in {French and English}: The role of
  syntactic hierarchy.
\newblock {\em Language and Cognitive Processes\/} 17(4):371--404.

\bibitem[{Grefenstette et~al.(2015)Grefenstette, Hermann, Suleyman, and
  Blunsom}]{grefenstette2015learning}
Edward Grefenstette, Karl~Moritz Hermann, Mustafa Suleyman, and Phil Blunsom.
  2015.
\newblock Learning to transduce with unbounded memory.
\newblock In {\em {Advances in Neural Information Processing Systems 28}\/}.
  pages 1828--1836.

\bibitem[{Hashimoto et~al.(2016)Hashimoto, Xiong, Tsuruoka, and
  Socher}]{hashimoto2016joint}
Kazuma Hashimoto, Caiming Xiong, Yoshimasa Tsuruoka, and Richard Socher. 2016.
\newblock A joint many-task model: Growing a neural network for multiple {NLP}
  tasks.
\newblock In {\em NIPS 2016 Continual Learning and Deep Networks Workshop\/}.

\bibitem[{Hockenmaier and Steedman(2007)}]{hockenmaier2007ccgbank}
Julia Hockenmaier and Mark Steedman. 2007.
\newblock {CCGbank}: {A} corpus of {CCG} derivations and dependency structures
  extracted from the {Penn Treebank}.
\newblock {\em Computational Linguistics\/} 33(3):355--396.

\bibitem[{Jozefowicz et~al.(2016)Jozefowicz, Vinyals, Schuster, Shazeer, and
  Wu}]{jozefowicz2016exploring}
Rafal Jozefowicz, Oriol Vinyals, Mike Schuster, Noam Shazeer, and Yonghui Wu.
  2016.
\newblock Exploring the limits of language modeling.
\newblock {\em arXiv preprint arXiv:1602.02410\/} .

\bibitem[{Lau et~al.(2016)Lau, Clark, and Lappin}]{lau2016grammaticality}
Jey~Han Lau, Alexander Clark, and Shalom Lappin. 2016.
\newblock Grammaticality, acceptability, and probability: A probabilistic view
  of linguistic knowledge.
\newblock {\em Cognitive Science\/} .

\bibitem[{Le~Godais et~al.(2017)Le~Godais, Linzen, and
  Dupoux}]{legodais2017comparing}
Ga\"{e}l Le~Godais, Tal Linzen, and Emmanuel Dupoux. 2017.
\newblock Comparing character-level neural language models using a lexical
  decision task.
\newblock In {\em Proceedings of the 15th Conference of the European Chapter of
  the Association for Computational Linguistics: Volume 2, Short Papers\/}.
  Association for Computational Linguistics, Valencia, Spain, pages 125--130.

\bibitem[{Lewis et~al.(2016)Lewis, Lee, and Zettlemoyer}]{lewis2016lstm}
Mike Lewis, Kenton Lee, and Luke Zettlemoyer. 2016.
\newblock {LSTM} {CCG} parsing.
\newblock In {\em Proceedings of the 2016 Conference of the North American
  Chapter of the Association for Computational Linguistics: Human Language
  Technologies\/}. pages 221--231.

\bibitem[{Linzen et~al.(2016)Linzen, Dupoux, and
  Goldberg}]{linzen2016assessing}
Tal Linzen, Emmanuel Dupoux, and Yoav Goldberg. 2016.
\newblock Assessing the ability of {LSTMs} to learn syntax-sensitive
  dependencies.
\newblock {\em Transactions of the Association for Computational Linguistics\/}
  4:521--535.

\bibitem[{Linzen et~al.(2017)Linzen, Goldberg, and
  Dupoux}]{linzen2017agreement}
Tal Linzen, Yoav Goldberg, and Emmanuel Dupoux. 2017.
\newblock Agreement attraction errors in neural networks.
\newblock In {\em Proceedings of the CUNY Conference on Human Sentence
  Processing\/}.

\bibitem[{Marcus et~al.(1993)Marcus, Marcinkiewicz, and
  Santorini}]{marcus1993building}
Mitchell~P. Marcus, Mary~Ann Marcinkiewicz, and Beatrice Santorini. 1993.
\newblock Building a large annotated corpus of {English}: The {Penn}
  {Treebank}.
\newblock {\em Computational Linguistics\/} 19(2):313--330.

\bibitem[{{Mart{\'\i}nez Alonso} and Plank(2017)}]{alonso2016multitask}
H{\'e}ctor {Mart{\'\i}nez Alonso} and Barbara Plank. 2017.
\newblock When is multitask learning effective? {Semantic} sequence prediction
  under varying data conditions.
\newblock In {\em Proceedings of the Conference of the European Chapter of the
  Association for Computationl Linguistics\/}.

\bibitem[{Mikolov et~al.(2010)Mikolov, Karafi{\'a}t, Burget, Cernock{\`y}, and
  Khudanpur}]{mikolov2010recurrent}
Tomas Mikolov, Martin Karafi{\'a}t, Lukas Burget, Jan Cernock{\`y}, and Sanjeev
  Khudanpur. 2010.
\newblock Recurrent neural network based language model.
\newblock In {\em Proceedings of Interspeech\/}.

\bibitem[{Socher(2014)}]{socher2014recursive}
Richard Socher. 2014.
\newblock {\em Recursive Deep Learning for Natural Language Processing and
  Computer Vision\/}.
\newblock Ph.D. thesis, Stanford University.

\bibitem[{S{\o}gaard and Goldberg(2016)}]{sogaard2016deep}
Anders S{\o}gaard and Yoav Goldberg. 2016.
\newblock Deep multi-task learning with low level tasks supervised at lower
  layers.
\newblock In {\em Proceedings of the 54th Annual Meeting of the Association for
  Computational Linguistics (Volume 2: Short Papers)\/}. Association for
  Computational Linguistics, Berlin, Germany, pages 231--235.

\bibitem[{Staub(2009)}]{staub2009interpretation}
Adrian Staub. 2009.
\newblock On the interpretation of the number attraction effect: Response time
  evidence.
\newblock {\em Journal of Memory and Language\/} 60(2):308--327.

\bibitem[{Steedman(2000)}]{steedman2000syntactic}
Mark Steedman. 2000.
\newblock {\em The syntactic process\/}.
\newblock MIT Press.

\bibitem[{Sundermeyer et~al.(2012)Sundermeyer, Schl{\"u}ter, and
  Ney}]{sundermeyer2012lstm}
Martin Sundermeyer, Ralf Schl{\"u}ter, and Hermann Ney. 2012.
\newblock {LSTM} neural networks for language modeling.
\newblock In {\em {Proceedings of the 13th Annual Conference of the
  International Speech Communication Association (INTERSPEECH)}\/}. pages
  194--197.

\bibitem[{{Theano Development Team}(2016)}]{theano}
{Theano Development Team}. 2016.
\newblock \href{http://arxiv.org/abs/1605.02688}{{Theano: A {Python} framework
  for fast computation of mathematical expressions}}.
\newblock {\em arXiv e-prints\/} abs/1605.02688.
\newblock
  \href{http://arxiv.org/abs/1605.02688}{http://arxiv.org/abs/1605.02688}.

\bibitem[{Vaswani et~al.(2016)Vaswani, Bisk, Sagae, and
  Musa}]{vaswani2016supertagging}
Ashish Vaswani, Yonatan Bisk, Kenji Sagae, and Ryan Musa. 2016.
\newblock Supertagging with {LSTMs}.
\newblock In {\em Proceedings of NAACL-HLT\/}. pages 232--237.

\bibitem[{Wagers et~al.(2009)Wagers, Lau, and Phillips}]{wagers2009agreement}
Matthew~W. Wagers, Ellen~F. Lau, and Colin Phillips. 2009.
\newblock Agreement attraction in comprehension: Representations and processes.
\newblock {\em Journal of Memory and Language\/} 61(2):206--237.

\bibitem[{Xu et~al.(2015)Xu, Auli, and Clark}]{xu2015ccg}
Wenduan Xu, Michael Auli, and Stephen Clark. 2015.
\newblock {CCG} supertagging with a recurrent neural network.
\newblock In {\em Proceedings of the 53rd Annual Meeting of the Association for
  Computational Linguistics and the 7th International Joint Conference on
  Natural Language Processing (Volume 2: Short Papers)\/}. Association for
  Computational Linguistics, Beijing, China, pages 250--255.

\end{thebibliography}
\appendix

\clearpage
\newpage
\section{Appendix}

\begin{figure}[b]
\centering

\subfloat[][\newcite{bock1992regulating}: PS]{
    \includegraphics[width=0.95\linewidth]{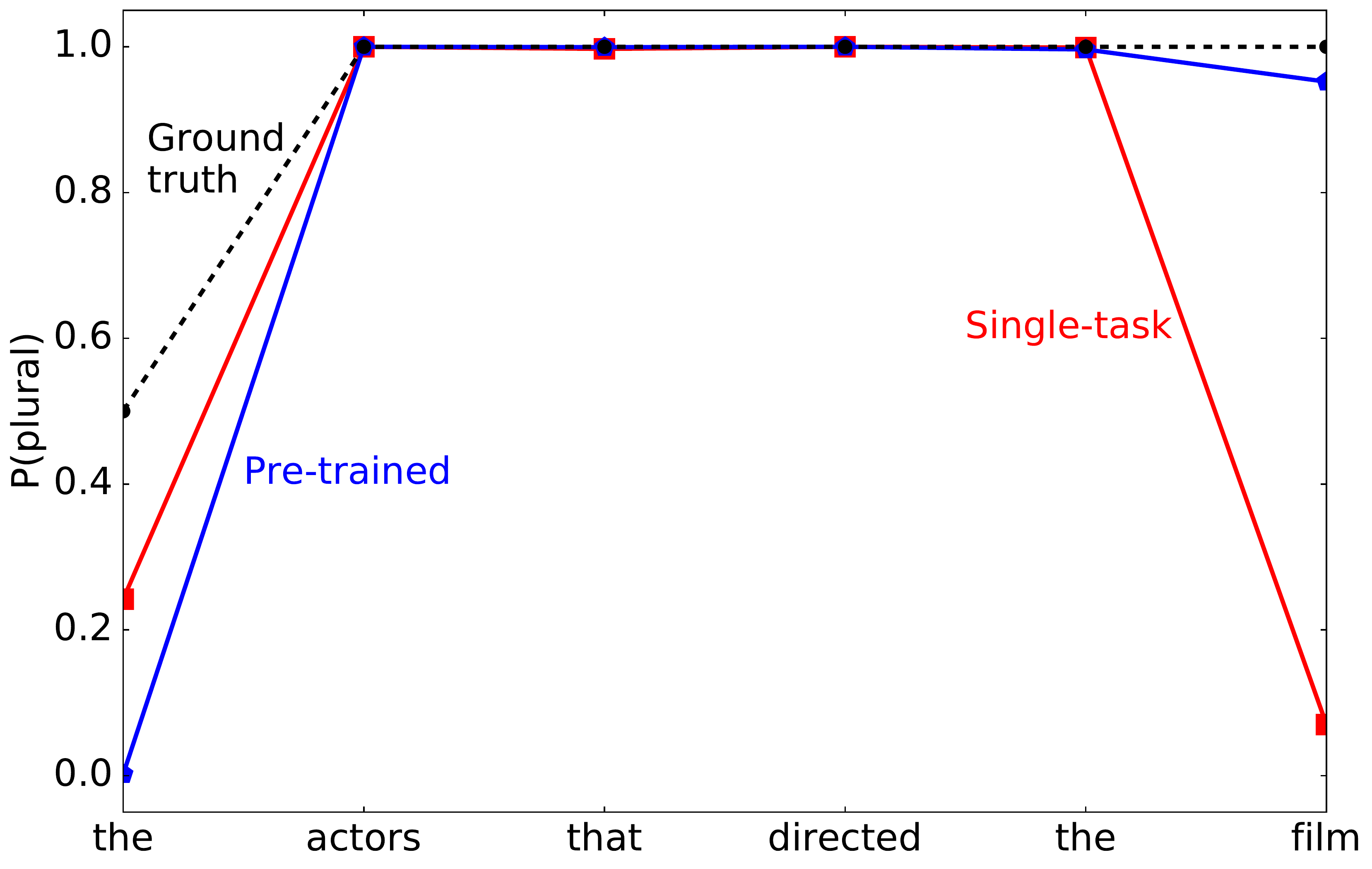}
    \label{fig:bock_example}
}

\subfloat[][\newcite{wagers2009agreement}: PS]{
	\includegraphics[width=0.95\linewidth]{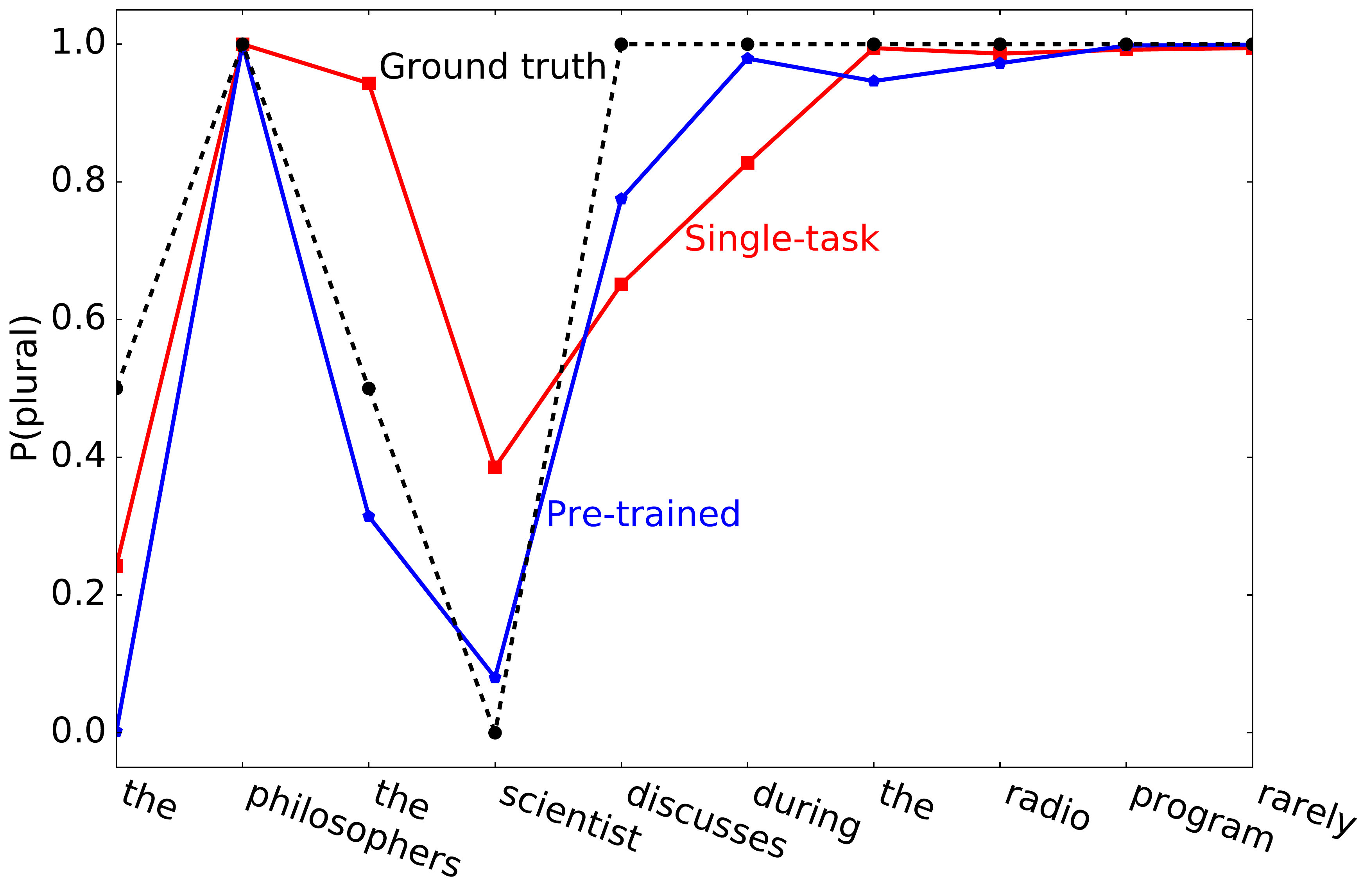}
    \label{fig:wagers_example_ps}
}

\subfloat[][\newcite{wagers2009agreement}: SP]{
	\includegraphics[width=0.95\linewidth]{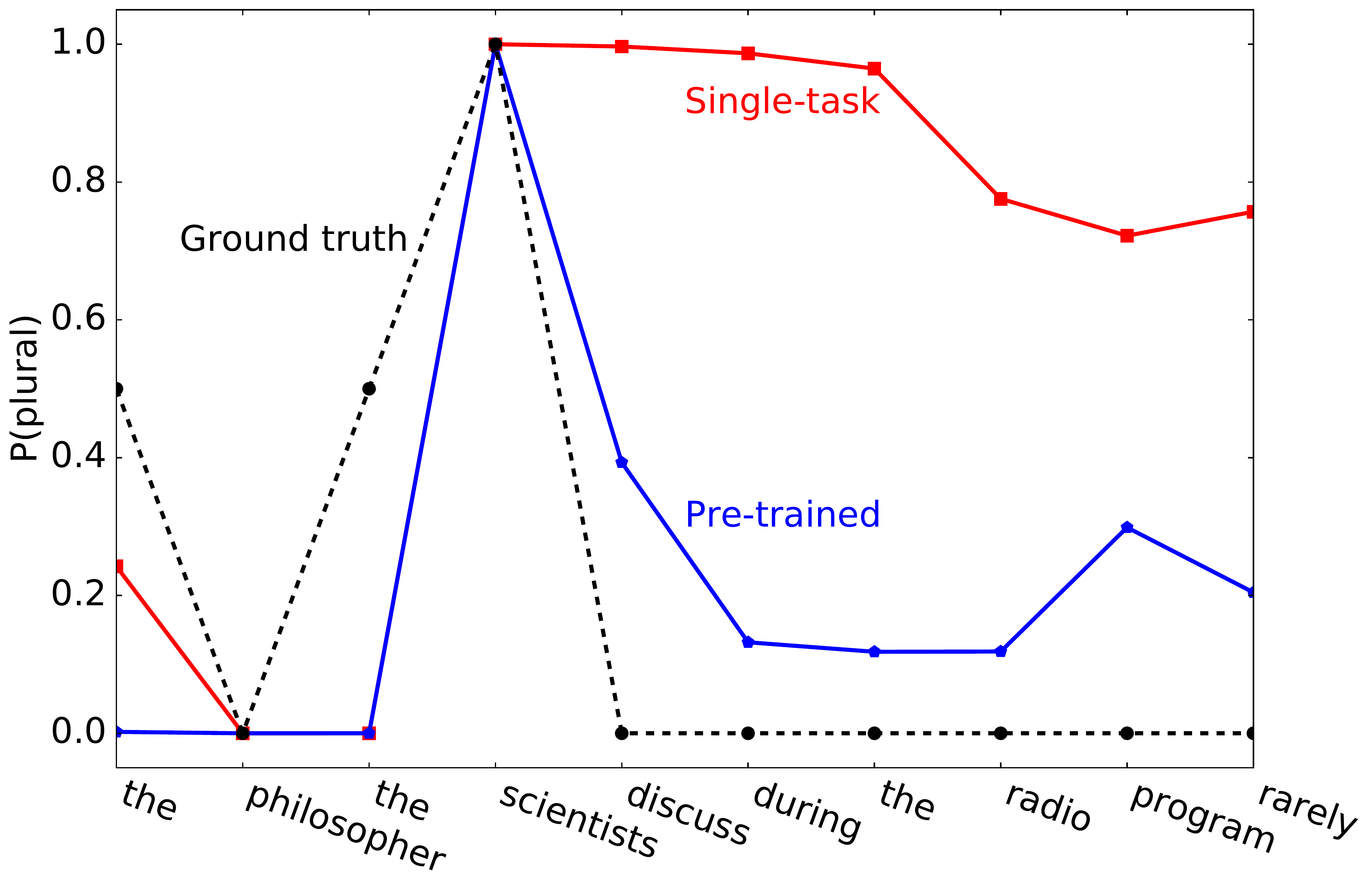}
    \label{fig:wagers_example_sp}
}

\caption{Probability of a plural prediction after each word in the sentence for three sample sentences. The black dotted line indicates the grammatical ground truth.}
\label{fig:psych_examples}
\end{figure}

\begin{figure*}[bt]
\centering
\subfloat[][Unit 30: approximately tracks the number of the\\currently relevant subject]{
    \includegraphics[width=0.45\linewidth]{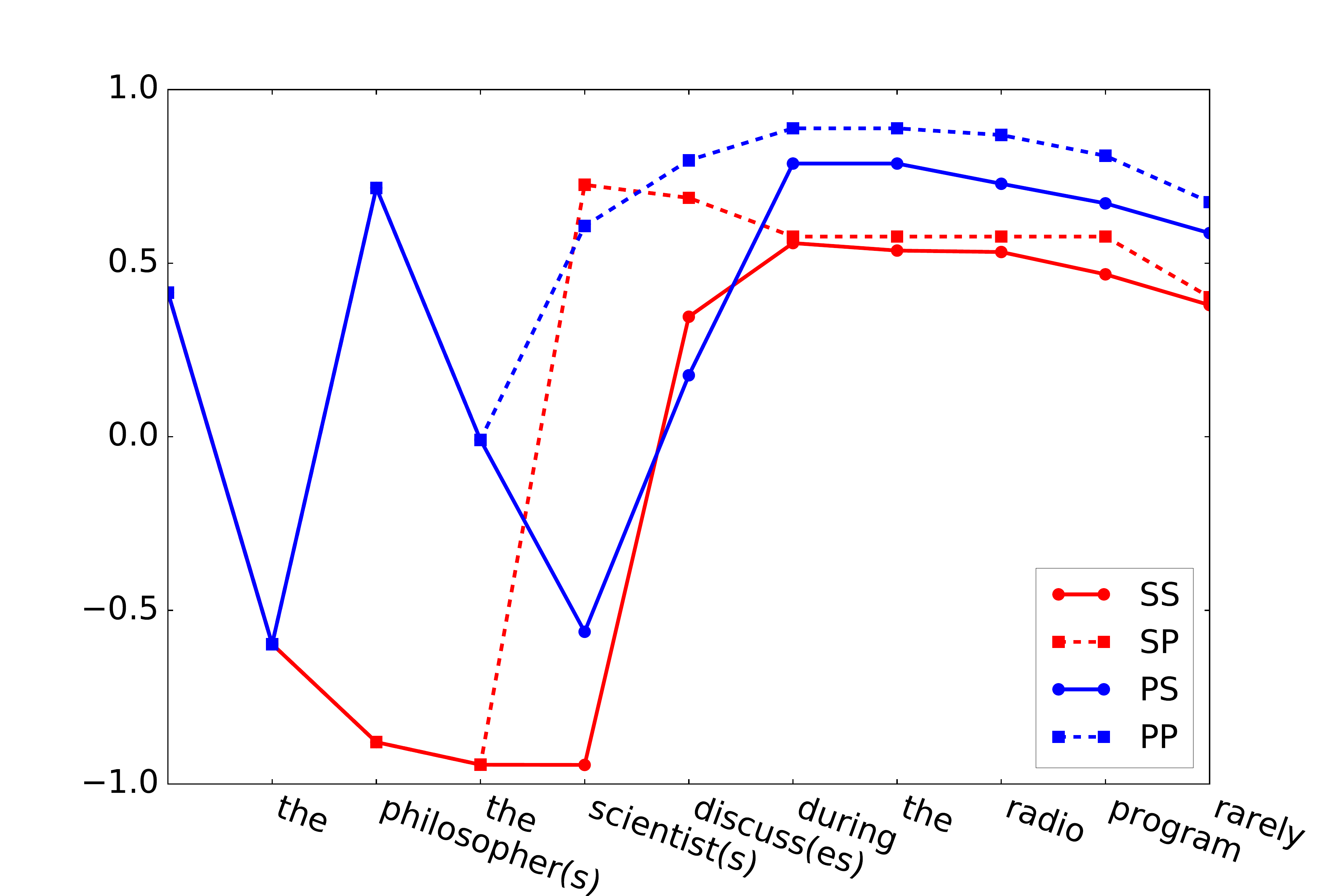}
}
\subfloat[][Unit 50: only active within noun phrases]{
    \includegraphics[width=0.45\linewidth]{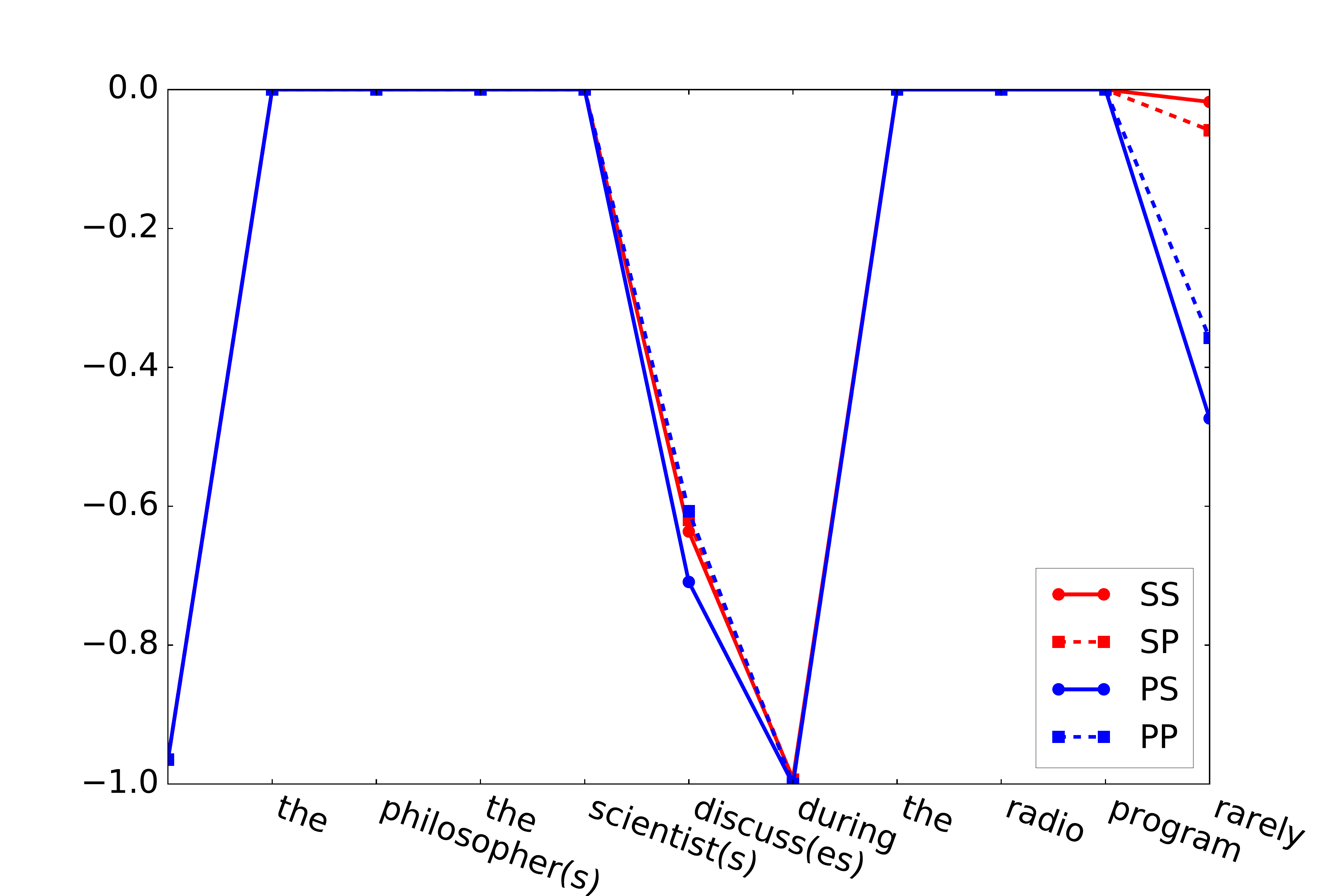}
}

\subfloat[][Unit~73: represents of the number of the main\\clause subject]{
    \includegraphics[width=0.45\linewidth]{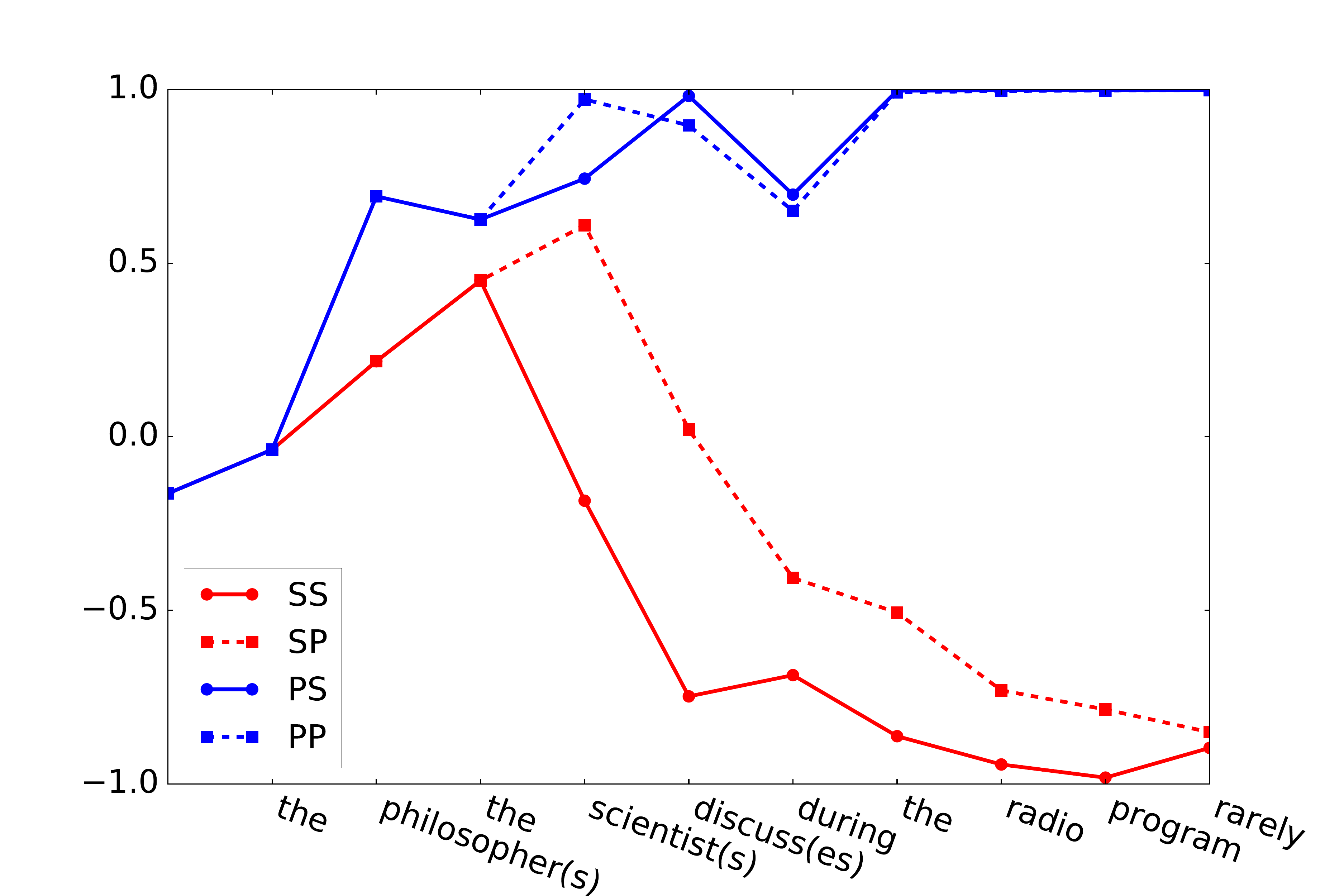}
}
\subfloat[][Unit~86: approximately tracks the number of the\\currently relevant subject)]{
    \includegraphics[width=0.45\linewidth]{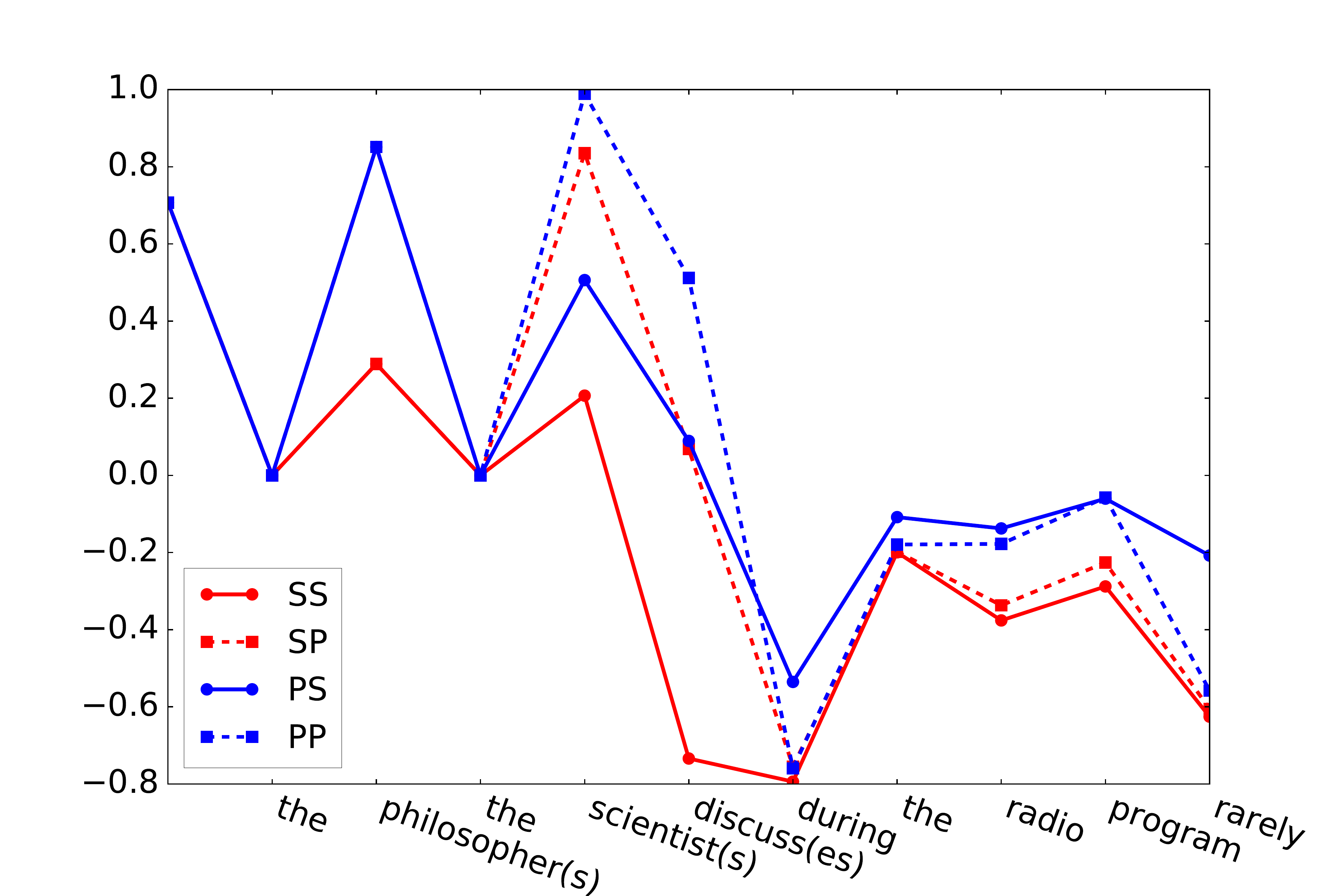}
}
\caption{Activations of a sample of interpretable units throughout an example sentence from \newcite{wagers2009agreement}, for all four number configurations.}
\label{fig:activations}
\end{figure*}

This appendix presents figures based on sentences with relative clause (see Section \ref{sec:relative_clauses}).  Figure \ref{fig:psych_examples} tracks the word-by-word predictions that the single-task model and the pre-trained model make for three sample sentences; the grammatical ground truth is indicated with a dotted black line. Overall, the pre-trained model is closer to the ground truth than the single-task model, even in cases where both models ultimately make the correct prediction (Figure~\ref{fig:wagers_example_ps}). Figures~\ref{fig:bock_example} and \ref{fig:wagers_example_sp} show cases in which an attractor in an embedded clause misleads the single-task but not the pre-trained one. Finally, Figure~\ref{fig:activations} shows a sample of four units that appear to track interpretable aspects of the sentence.

\clearpage

\end{document}